# Development of a Fuzzy Expert System based Liveliness Detection Scheme for Biometric Authentication

Avinash Kumar Singh*, Piyush Joshi, G C Nandi

*Robotics and Artificial Intelligence Laboratory, Indian Institute of Information Technology, Allahabad, India-211012*

**Abstract**

Liveliness detection acts as a safe guard against spoofing attacks. Most of the researchers used vision based techniques to detect liveliness of the user, but they are highly sensitive to illumination effects. Therefore it is very hard to design a system, which will work robustly under all circumstances. Literature shows that most of the research utilize eye blink or mouth movement to detect the liveliness, while the other group used face texture to distinguish between real and imposter. The classification results of all these approaches decreases drastically in variable light conditions. Hence in this paper we are introducing fuzzy expert system which is sufficient enough to handle most of the cases comes in real time. We have used two testing parameters, (a) under bad illumination and (b) less movement in eyes and mouth in case of real user to evaluate the performance of the system. The system is behaving well in all, while in first case its False Rejection Rate (FRR) is 0.28, and in second case its FRR is 0.4.
© 2014 The Authors. Published by Elsevier B.V.

## 1. Main text

Liveliness detection is a way to detect that the person is live or not while submitting biometric trait for verification, in order to ensure that only the authorized person is using the system. There could be other's ways to prevent spoofing attacks suggested by Schuckers [3], such as combining biometric trait with a pin or a smart card, supervising the verification process, using multi-modal biometrics, and checking liveliness of the user. Among all liveliness detection is the most reliable and user friendly way to prevent faces spoofing attacks [7].

* Corresponding author. Tel.: +919005722861;
*E-mail address:* avinashkumarsingh1986@gmail.com. Online available at: http://searchdl.org/index.php/book_series/view/941

Face spoofing is an attack where attacker tries to make fool to the authentication system by using various artifacts of the enrolled user [20]. Artifacts could be anything like photo, video, or mask. The problem lays the way face recognition works briefly described in Figure 1. Throughout the process, face recognition system doesn't care about the person liveliness, it just needs face whether it is real or imposter system doesn't bother about it [18][19]. Attackers exploit this limitation of the face recognition system just by placing photo, or video of the enrolled user, and easily bypass the security mechanism. Researchers observed the need of security mechanism which can ensure reliability of the system and proposed various ways to deal with this problem. On the basis of literature we have grouped possible solution in three main categories (1) liveliness detection by using challenge and response method [17] (2) liveliness detection by utilizing face texture (image quality), and (3) liveliness detection by combining two or more biometrics (multi-modal approach). In challenge and response (basically used to distinguish photo from real user) method system throws some challenge in terms of eyes and mouth movement which can only be performed by real user not by photo, and analyzes the response in account of the given challenges. In challenge and response most of the researchers [4][5][6] have utilized eye blinking, while others exploits image quality (texture, edges, etc.) information to distinguish between real and imposter. Researchers have used local binary pattern (LBP) [9], Sparse Low Rank Bilinear Logistic Regression [8], low-level feature descriptors [10], etc.to evaluate the quality of image. Multimodal approach mostly uses speech and face as the combination to deal with this attack. In this regard research [11][12][13] have utilized Face-voice fusion, Mouth-motion and speech, Face voice correlation, etc. mechanism to prevent spoofing attacks.

Each approach has its own pros and cons, in challenge and response user must have to satisfy all the challenges, otherwise he/she will be treated as imposter, which is very hard to achieve when the environment is not static. This can increase the false rejection rate (FRR). Face texture is a good way, it can discriminate on the basis of surface roughness (face has more roughness while the medium which attacker uses is smooth (glossy, reflective in nature)). In multimodal biometrics, permission is granted on behalf of the resultant score. The problem occurs, when one fake biometric trait results in higher acceptance rate against actual biometric trait, which equalizes the overall score and results in increment of False Acceptance Rate. All these approaches has hardness in their core, which is not sufficient enough to handle real life scenarios, hence in this paper we are proposing fuzzy logy concept to deal with this problem, we have fuzzified the input variables (eye, mouth movement and image quality) and written some rules for classification. The rest of the paper is structured as follows: Section 2 describes robust framework for detecting the liveliness of the person with experimental setup that we have used in our approach followed by Section 3 that coherently states the results obtained and their discussion. In Section 4, we conclude the paper with its contribution towards the face biometrics and its future prospects.

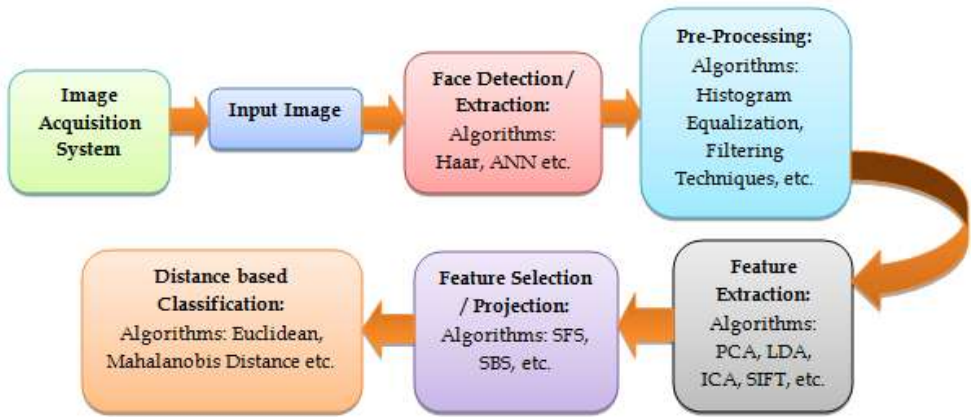

Figure 1. Statistical Way of Face Recognition

## 2. Proposed Framework and Experimental Setup

We are using Mamdami style inference system proposed by Professor Ebrahim Mamdani [2]. For better explanation, we have divided our framework into five different modules discussed below:

*2.1. Inputs to the System*

The proposed system takes three parameters as input eye movement, mouth movement and image quality. We are using Haar cascade classifier devised by Viola and Jones [15] to detect eye and mouth in the face, and for detecting the movement inside them we are using Scale Invariant Feature Transformation (SIFT) proposed by David [14]. SIFT is used for extracting distinct features of eyes and mouths in each frame. These features are invariant to image scaling, translation, and rotation, and partially invariant to illumination changes. Hence if there is any movement in mouths and eyes, these features will get change, resultant highly mismatches to the next frame0s mouths and eye features shown in Figure3.

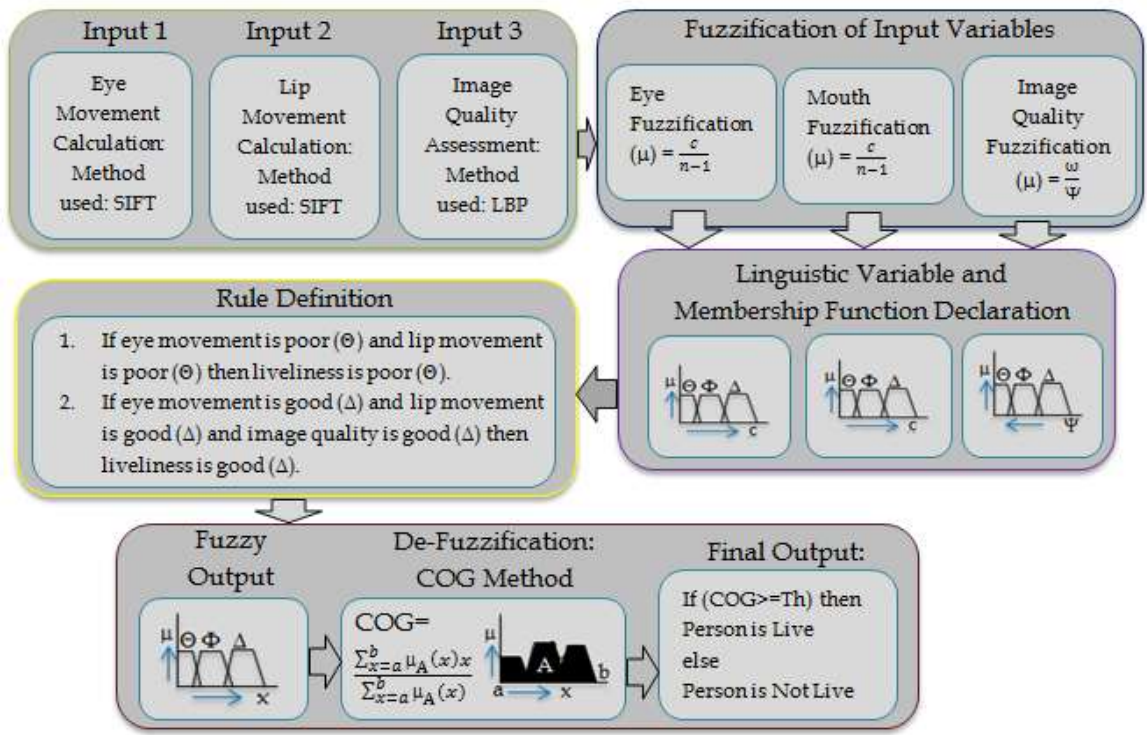

Figure 2. Fuzzy Expert System to Prevent Spoofing Attack

This could reflect movement in these parts. Image quality is measured by measuring surface texture of the face by using Local Binary Pattern (LBP) introduced by [9]. LBP is used to code each pixel information based on their neighbors locally (Here we used 8 neighbors, with radius (R=1)), described in expression (1). If the intensity ($i_c$) of the center pixel ($x_c, y_c$) is less than its neighbor's intensity ($i_p$), then its neighbor's ($x_p, y_p$) value will be assumed 1 else 0.

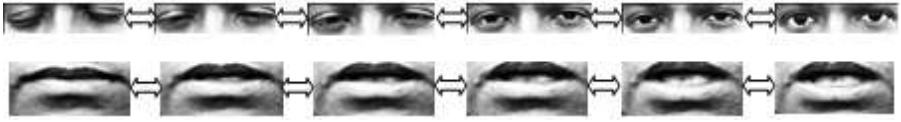

Figure 3. Eye and Mouth Movement in Face

$$LBP(x_c, y_c) = \sum_{i=0}^{p-1} 2^p \times S(i_p - i_c), \text{ where } S(x) = \begin{cases} 1 & if\ x \geq 0 \\ 0 & therwise \end{cases}$$

This representation basically denotes homogeneity (pixel having same intensity) in the image, which shows surface roughness. The medium which attacker uses is mostly glossy in nature (Mobile/Laptop used for video imposter attack), having some reflectance property resultant homogeneity in captured image, while real faces having less reflectance due to 3D face structure and roughness. This difference can be easily visualized in Figure4, white circle shows the window where homogeneity is more in comparison to real person.

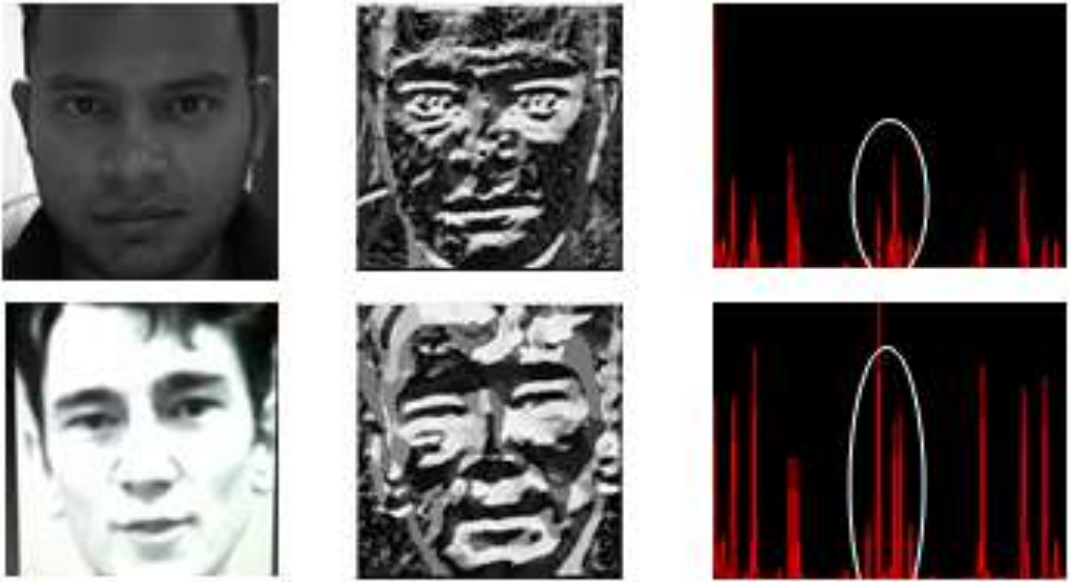

Figure 4. Real and Fake Faces: First row shows the real person photo, face texture and histogram plot (Y axis denotes frequency, while x axis denotes intensities values ranging from 0-255). Second row represents the fake face with face texture and histogram plot

*2.2. Fuzzification* Method, Linguistic Variable and Membership Function used

1. *Eye & Mouth movement fuzzification ($\mu$):*
   $\mu = c/((n-1))$, Where c is the counter used for counting movement in eye and mouth. "n" is the number of frames used for analyzing the movement in eyes and mouths.
2. *Image Quality fuzzification ($\mu$):*

   $\mu = \begin{cases} \frac{256}{\mu} & if\ \Psi \geq 1 \\ 1 & otherwise\ 1 \end{cases}$  and $\sum_{i=k}^{l} F_i$

Where (Ψ) is the summation of homogeneous region decided by analyzing the histogram of fake and real faces, k and l denotes the starting and ending position while F shows the frequency of each pixel. We have used three linguistic variable as said earlier eye, mouth movement, and image quality and specified their values as poor (Θ), average (Φ), and good (Δ) respectively. Range of these variables is summarized in table 1. Membership function used here is Trapezoidal [1].

Table 1. Fuzzification of input variables

| Linguistic Variables | Range (Interval Selection) | | |
|---|---|---|---|
| | Poor(Θ) | Average(Φ) | Good(Δ) |
| Eye Movement | [0-6] | [5-11] | [10-19] |
| Mouth Movement | [0-5] | [4-10] | [10-19] |
| Image Quality | [900-1300] | [800-600] | [500-256] |
| Output | [0-0.4] | [0.3-0.6] | [0.5-1] |

*2.3. Rule definition and Defuzzification*

We can define k rules: where $\prod_{i=1}^{n} \Omega_i$ where, n is the number of input linguistic variables, and Ω denotes the possible linguistic values of that. But for the representation purpose we are showing some of the rules that we used. We can use AND/OR operators as a conjunction, where AND represents the maximum between two while OR represents the minimum. Out of k rules m rules are fired according to the match shown in example. Rules should be written in such a way that there should not be redundancy between them and they should reflect what you have thought for. Glimpse of these rules are:

1. If eye movement is poor AND mouth movement is poor then output is poor.
2. If eye movement is good OR mouth movement is good AND image quality is good then output is good.
3. If eye movement is good OR mouth movement is poor AND image quality is good then output is good.

a) *Execution of rules*

Let we have given values, (a) eye movement: c = 15, n=20 (b) mouth movement: c=10 and n=20 and (c) face texture value = 400 then by fuzzifying all there we will get their membership value as $\Psi_A(x)$ = 0.79, $\Psi_B(x)$ =0.53, $\Psi_C(x)$ =0.64, according to these values we can assign linguistic terms to them, like eye movement is good, mouth movement is good, and Image quality is good, these linguistic values will trigger rule 2. The output will be: min ((max ($\Psi_A(x)$ =0.79, $\Psi_B(x)$ =0.53 )), $\Psi_C(x)$ =0.64) = [ $\Psi_O(x)$ =0.64]. This will fall into the output linguistic value good with the membership value: 0.64.

b) *Defuzzification*

Through aggregation we are adding all the regions came from the previous step shown in proposed framework Figure 2, in order to analyze the output. But for making the decision from the system we have to have crisp value. Defuzzification helps to get that, there are several methods to perform

Defuzzification mentioned in the literature, but the most widely used is Center of Gravity shown in expression2.

$$COG = \frac{\sum_{x=a}^{b} \mu_O(x) \times x}{\sum_{x=a}^{b} \mu_O(x)}$$

Where a and b shows the starting and ending of the aggregate region, $\mu_O(x)$ represents the membership of the output, while x is the base, fragmented over the fix distance in order to calculate COG.

## 3. Results and Discussions

We have rigorously tested the proposed framework by using 3 different attacks medium such as (a) photo imposter attack (medium used laptop), (b) photo imposter attack (medium used 2D plain paper) (c) High Definition video imposter attack (medium used mobile) together with real person. We have used 95 fake faces from University of Essex, UK (face94) [16] in order to create test set (a), while 25 real faces from Robotics and Artificial Intelligence Laboratory face database, the same database of real users is used for creating the high definition video imposter attack as well as 2D plain paper photo imposter attack. We have chosen different mediums because each one has its own different reactance power. Apart from these we have created another 3 sets for testing the robustness of the system (a) Testing parameter-I (ideal condition): Normal daily life condition where visibility is clear, and user is showing significant movement in eyes and mouth when comes for authentication (b) Testing Parameter-II (under bad illumination) and (c) Testing Parameter-III (no or very less movements in eyes and mouth with respect to real user). Results under these circumstances are summarized in Table2.

Table 2. Efficiency of the proposed framework

| Attack with medium used for testing | Accuracy (%) | | |
|---|---|---|---|
| | Ideal Case | Testing Parameter-I | Testing parameter-II |
| Real Person | 100 | 72 | 60 |
| Photo Attack with Laptop | 100 | 100 | 100 |
| Photo Attack with 2D plain paper | 100 | 100 | 100 |
| HD video attack with HD mobile phone | 100 | 100 | 100 |

In ideal conditions system is showing 100 % classification in all, while in bad illumination its efficiency decreases, and we are getting False Rejection Rate as 0.28. This is also because in bad illumination system is not able to detect eye and mouth in face, hence not able to capture movements inside them, due to this texture quality is also not good. Technically we can say that linguistic values of eyes and mouth movements map to either poor or average. Image quality also map to this region based on the person's face fairness, hence resultant output is either poor or average. For testing under testing parameter-III, we have divided the real user into three groups (a) includes poor movements, (b) average movements and (c) good movements. On the basis of these test sets we are getting FRR as 0.4. In order to reduce FRR of testing parameter -III, user has to give just a little attention to the system (like a smile a blink etc.), which is very simple. All these results are graphically represented in Figure 5. The flexibility of the proposed framework is, we can use existing literature over this and can get significant improvement in classification.

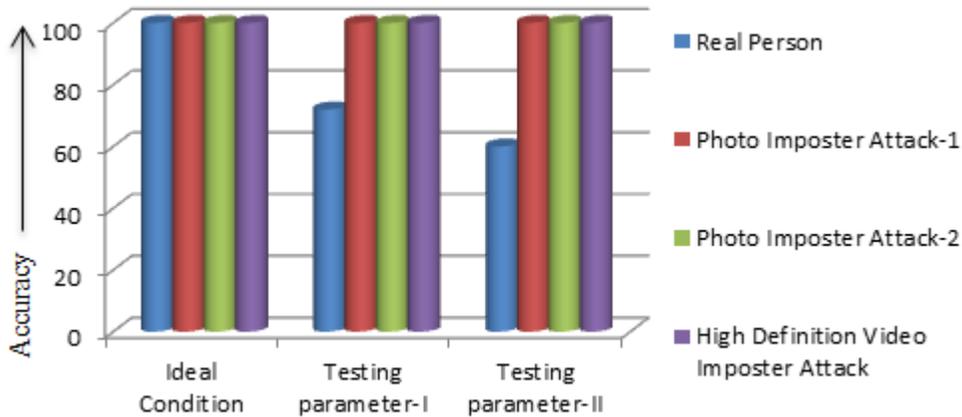

Figure 5. Result analysis under different attack parameters

## 4. Conclusion

Proposed system is able to capture real world scenario in a better way than the classical approach of hard computing. Experimental results show the efficiency of proposed framework, and strengthen this new way of classification. System is showing a perfect classification in ideal case while in testing parameter-I and II its FRR is 0.28 and 0.4 respectively, in case of all three attack its accuracy is very good with zero FAR. We are getting miss classification also because, we have defined our rule base system in such a way that whenever it sense some attacks it will match it to its respective class. It could be resolved just by giving a slight attention to the system in these two testing parameters. Efficiency of the system depends on several parameters like (a) how you are fuzzifying the input parameter, (b) range selection for linguistic values, (c) selection of membership function and most importantly (d) how you are setting up your rule database. These parameters could be assumed as tuning parameter of the system, in order to get best performance these should be tuned properly. While for selecting the threshold (after Defuzzification) to discriminate between real and imposter, we can use Equal Error Rate (EER).

**Acknowledgements**

Author would like to thank all the students, research scholars, and project associates of Robotics and Artificial Intelligence lab of Indian Institute of Information Technology Allahabad for donating their face biometric data, and their cooperation during several time testing over this project.